\def\year{2018}\relax
\documentclass[letterpaper]{article} %DO NOT CHANGE THIS
\usepackage{aaai18}  %Required
\usepackage{times}  %Required
\usepackage{helvet}  %Required
\usepackage{courier}  %Required
\usepackage{url}  %Required
\usepackage{graphicx}  %Required
\usepackage{latexsym}
\usepackage{caption}
\usepackage{mwe}
\usepackage{tabularx}
\usepackage{xcolor}
\usepackage{multirow}
\usepackage{microtype}
\usepackage{lipsum}
\usepackage{epstopdf}
\usepackage{ragged2e}
\usepackage{subfigure}
\usepackage{array}
\usepackage{url}
\usepackage{booktabs}
\usepackage{relsize}
\usepackage[shortlabels]{enumitem}
\usepackage{amsmath}
\usepackage{makecell}
\setlist[enumerate]{noitemsep}

\makeatletter
\newcommand{\@BIBLABEL}{\@emptybiblabel}
\newcommand{\@emptybiblabel}[1]{}
\makeatother

\begin{document}
\title{Towards Building Large Scale Multimodal Domain-Aware Conversation Systems}

\author{ \qquad~~ Amrita Saha$^{1,2}$ ~\qquad\qquad~~ Mitesh M. Khapra$^2$ \qquad\qquad Karthik Sankaranarayanan$^1$\\
amrsaha4@in.ibm.com \qquad\qquad\qquad miteshk@cse.iitm.ac.in \qquad\qquad\qquad\qquad kartsank@in.ibm.com \\\\
    $^1$\Large{IBM Research AI} \qquad
    $^2$\Large{I.I.T. Madras, India}\\%
}

\maketitle
\begin{abstract}
While multimodal conversation agents are gaining importance in several domains such as retail, travel etc., deep learning research in this area has been limited primarily due to the lack of availability of large-scale, open chatlogs. To overcome this bottleneck, in this paper we introduce the task of multimodal, domain-aware conversations, and propose the MMD benchmark dataset. This dataset was gathered by working in close coordination with large number of domain experts in the retail domain. These experts suggested various conversations flows and dialog states which are typically seen in multimodal conversations in the fashion domain. Keeping these flows and states in mind, we created a dataset consisting of over 150K conversation sessions between shoppers and sales agents, with the help of in-house annotators using a semi-automated manually intense iterative process. 
With this dataset, we propose 5 new sub-tasks for multimodal conversations along with their evaluation methodology. We also propose two multimodal neural models in the encode-attend-decode paradigm and demonstrate their performance on two of the sub-tasks, namely text response generation and best image response selection. These experiments serve to establish baseline performance and open new research directions for each of these sub-tasks. Further, for each of the sub-tasks, we present a `per-state evaluation' of 9 most significant dialog states, which would enable more focused research into understanding the challenges and complexities involved in each of these states.\end{abstract}

\section{Introduction}

The recent progress with deep learning techniques for problems at the intersection of NLP and Computer Vision such as image captioning \cite{cap2}, video description \cite{DBLP:conf/cvpr/YuWHYX16}, image question answering \cite{vqa-visual-question-answering}, video question answering \cite{DBLP:journals/corr/ZengCCLNS16,DBLP:journals/corr/MaharajBCP16}, is owed largely due to the availability of large-scale open datasets for their respective tasks. However, even though there is a growing demand for chatbots that can converse using multiple modalities with humans in several domains such as retail, travel, entertainment, etc. the primary hindrance for deep learning research in this area has been the lack of large-scale open datasets. 
%that exemplify both of the challenges mentioned above. 
Though there has been recent work \cite{DBLP:journals/corr/SerbanSLCPCB16,DBLP:journals/corr/YaoPZW16,Serban:2016:BED:3016387.3016435} with different conversation datasets \cite{DBLP:conf/sigdial/LowePSP15,DBLP:journals/corr/VinyalsL15,DBLP:conf/naacl/RitterCD10}, the mode of interaction there is limited to text conversations only.
%rendering them inadequate for multimodal conversation research. 
While multimodal, human-to-human conversation transcripts (e.g. between shoppers and salespersons) might be available in industry settings, they are both limited in scale and proprietary, thus hindering open research.
\begin{figure*}[!ht]
  \includegraphics[width=\textwidth]{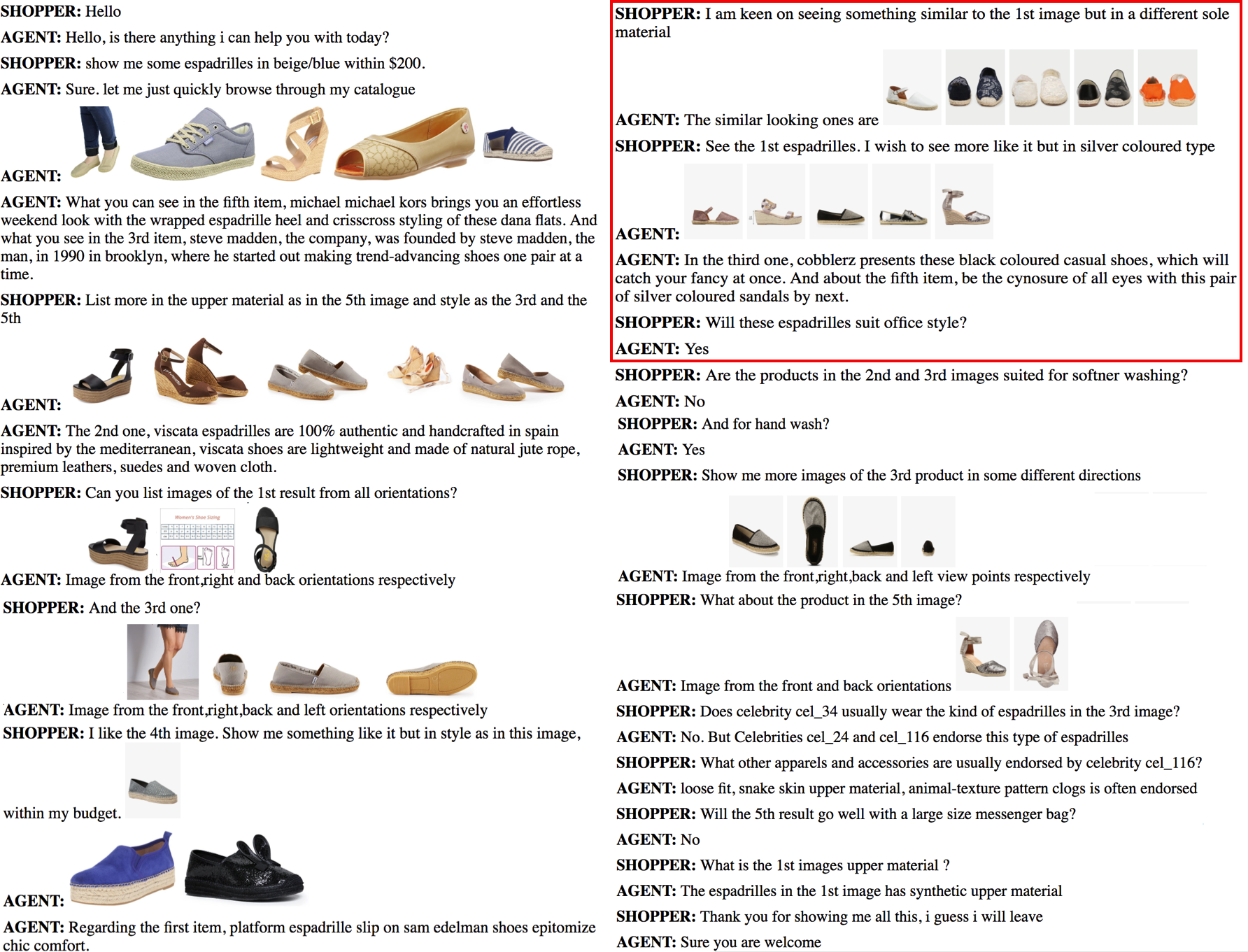}
   \caption{\label{fig:dialogue} Example Dialog session between a shopper and an agent}

\end{figure*}

\-\ \-\ In this paper, we attempt to alleviate these challenges by developing a large-scale multimodal conversational dataset in the retail domain that embodies the required generic capabilities for such autonomous agents. Since the actual transcripts are both limited and proprietary, we conducted a series of interviews with a large number of retail salespersons from the fashion domain and developed the dataset interactively with them in the following semi-automatic manner. The domain experts described in detail, various phases of the sales process which were materialized into 84 states for a conversational agent. Each such state had a specific intent (e.g. a shopper wanting to see more items similar to a specific one identified, or asking for a style tip or about the latest trends being endorsed by celebrities, etc). Corresponding to each such intent, a collection of representative utterance patterns involving both text and images were designed. Each of these states exhibited multimodality (i.e. involving text and images) in both the utterance/response as well as the dialog context. The domain experts described various possible sales flows where the customer went from one state to another, which were captured by transitions of an expert model automata between these states. The experts then inspected the outputs of controlled runs of the automata, provided feedback, which was thereafter employed to further refine the automata. Proceeding in this \emph{manually-intensive} and \emph{iterative} manner under the supervision of domain experts, we produced a large-scale dataset of over 150K multimodal dialogs capturing a wide variety of conversational sessions exhibiting the challenges described earlier. Note that such a data collection could not have been achieved with typical crowdsourcing platforms such as Amazon Mechanical Turk using ordinary crowd workers since it was necessary to be performed under the supervision of fashion sales experts exhibiting domain-specific understanding. 
\\
\-\ \-\ 

One might argue that such \textit{semi-simulated} conversations do not capture the flavor of \textit{natural} conversations. However, this is not entirely true for the conversations in our dataset. First, the constant supervision and inputs from domain experts ensures that the conversation remains grounded and is close to real world conversations in the fashion domain. Secondly, the actual utterances in the datset are not machine generated but solicited from in-house annotators. In particular, for any given state (say, ``express-interest-in-apparel''), in-house annotators were asked to provide natural language sentences corresponding to the state (for example, ``I think the second shirt looks great''). This is in contrast to the recent dialog datasets released by  \cite{DBLP:journals/corr/BordesW16} where the utterances are also machine generated. Also note that while there are 84 states in the conversation (which is a reasonable number given that the conversation is restricted to a specific domain), the fashion experts suggested various ways in which these states combine in natural conversations which results in a large number of paths that the conversation can follow. Again, since these paths were supervised by fashion experts, they are very similar to natural conversations in the domain. Collecting completely \textit{natural} real world conversation data at this scale is clearly infeasible and a data collection strategy of close-human machine interaction is perhaps a reasonable middle ground.

An example of a typical multimodal conversation in our dataset involving both text and images, and exhibiting domain knowledge is shown n Figure \ref{fig:dialogue}. Notice that the \textit{response} generated by the system (or agent) can either be text (for example, see the 3rd turn inside the highlighted portion in red) or a set of images (as seen in the 1st turn there). Similarly, at every point, the context available to the system is multimodal which the system has to process to generate a coherent response. For example, to give a relevant response in Turn 2 inside the box the agent needs to first pay attention to the words \textcolor{red}{``espadrille''}, \textcolor{red}{``silver''} and the \textcolor{red}{sentiment} which are mentioned in the text and then understand the style (\textcolor{blue}{``strapped ankle''}) from the image to fetch images for ``\textcolor{red}{silver espadrille} with \textcolor{blue}{strapped ankle''} in response to the given context.

The body of work most relevant to ours is Visual QA\cite{vqa-visual-question-answering} involving a single question and response, the work of \cite{DBLP:journals/corr/MostafazadehBDG17,DBLP:journals/corr/DasKGSYMPB16} involving a sequence of QA pairs with a single image forming a dialog, and the work of \cite{DBLP:journals/corr/VriesSCPLC16} which focuses on scene understanding and reasoning from a single image. There are a few key differences between these datasets and our work as demonstrated by the example above. First, in these cases, all questions in a sequence pertain only to a single image. Secondly, their responses are always textual. However, as is the case with natural conversations amongst humans, in our work, (i) there could be multiple images providing context, (ii) these context images could change across turns during the course of the conversation, and (iii) the response at each step can be text or image(s) or both. We would also like to mention that there is some recent work by \cite{DBLP:journals/corr/yuiha} wherein they use fashion data from instagram to analyze visual features of fashion images and correlate them with likes and comments on social media. Unlike our work they do not focus on multimodal conversations.

%One of the main contributions of this work is to fill this data void by creating a dataset containing multdimodal conversations from the fashion domain. To this end, we first crawled structured and unstructured data from various fashion websites. The structured data contained information about style, color, pattern, type of apparel, etc. The unstructured data on the other hand contained an informative description of the apparel. Figure 1, shows a snapshot of the structured and unstructured data for one such fashion product. We then teamed up with fashion experts who helped us refine the structured data to create a fashion ontology. The experts also provided various samples and insights of the kind of conversations that typically take place in the fashion domain. <Struggling to write the semi-automated part here... not sure how to  say it so that it doesn't look very artificial data>      

Finally, in this paper, we propose a baseline framework to model the agent's responses in such multimodal conversations. In particular, we propose hierarchical dialog models for the textual and image response as two separate tasks and empirically estimate the feasibility of these tasks. We also discuss limitations that open new directions for research into these and multiple other tasks enabled by this new dataset. The following summarizes the main contributions of this work:
\begin{enumerate}[$\bullet$,leftmargin=*,noitemsep,topsep=0pt]
\item We introduce a Multimodal Conversation task which is significantly distinct from the sequential Visual QA driven dialog tasks mentioned before
%introduced in \cite{DBLP:journals/corr/DasKGSYMPB16,DBLP:journals/corr/MostafazadehBDG17} and \cite{DBLP:journals/corr/VriesSCPLC16}
\item We introduce a large dataset for this task and define several research tasks for evaluating it
\item We propose baseline multimodal encoder decoder models for two such tasks and define appropriate metrics for evaluating these tasks
\end{enumerate}

%Humans find it very natural to describe and converse about images using natural language text. For example, even in visual domains (such as fashion, furniture etc) the search for products on online platforms is largely driven by natural language text. The more intuitive form of fashion search is through a conversation with a fashion agent/ domain expert and the entire conversation is completely dependent on understanding the multimodal context as well as knowledge of the domain. Though dialogue systems has been of steady interest in the NLP community and recently in the deep learning community, there has not been much work on multimodal dialogue systems. Especially there is no publicly available datasets on multimodal dialogue. 

%\section{DataSet Description}
\section{Multimodal Dialogs (MMD) Dataset}

As mentioned in the previous section, a key contribution of this paper is a large-scale dataset of 2-party dialogs that seamlessly employ multimodal data in their utterances and context and also demonstrate domain-specific knowledge in their series of interactions. Towards this goal, in this section, we first describe the methodology employed for collecting this dataset and then explain in detail the various sub-tasks exhibited by the dataset that open up new research problems.

\subsection{Data Collection Methodology}

%One of the main objectives of this paper is to introduce a dataset for multimodal dialogue (involving modalities like text and image). Since no such realistic dataset is publicly available we used a semi-automated way to simulate the dataset generation. One of the intuitive domains where such a multimodal dialogue is applicable is online fashion shopping. Fashion e-commerce is inherently visually driven and its a common understanding amongst domain experts that conversational agents would play a major role in making shopping more intuitive for the wide community of shoppers coming from different demographics etc. This motivates an interesting natural setting for a multimodal conversation relying on structured knowledge base as well as semi-structured or unstructured data in different modalities. 

The data collection done in close coordination with a team of 20 fashion experts, primarily consisted of two steps, (i) curation and representation of a large-scale domain knowledge, and (ii) developing a large collection of multimodal conversations, each consisting of a series of interactions employing this knowledge. We next proceed to describe these two steps in detail.

%With this understanding, the objective of our work has been to simulate a multimodal dialogue dataset for the fashion e-commerce shopping case. Though simulated, each step starting from crawling catalog information of real products from these e-commerce sites, to creating the dialogues, was interjected by a manual inspection and filtering step by domain experts. At the end, a survey was done over XXX such dialogues by domain experts to judge whether the simulated dialogues are realistic or not.
 
\subsubsection{Domain Knowledge Curation}
%\subsubsection{Domain-knowledge Curation from unstructured multimodal data}

Through our series of interviews with the domain experts, we observed that a lot of the complexity in a natural conversation in this domain comes from the background knowledge that both the expert agent and the shopper employ in their conversation. The expert's domain knowledge is multitude in nature, varying from knowledge about which attire goes well with which accessory, to which celebrity is presently endorsing which kind of fashion items, or what kind of look is better suited for which occasion. Therefore, the first step in our data collection process was to curate this domain knowledge from unstructured multimodal content on the web at scale and represent them in a machine consumable manner. This process involved a series of steps as enumerated below:

\noindent \textbf{1.} Crawling over 1 Million fashion items from the web  along with their available semi/un-structured information and associated image(s).\\ % as shown in Table \ref{tbl:catalog_example}\\
\textbf{2.} Parsing different types of domain knowledge from the free-text information, and curating them in a structured form after a round of manual inspection by domain experts\\
\-\ \-\ \textbf{2a.} Creating a hand-crafted taxonomy of the different types of fashion items. For example, \textit{man $>$ apparel $>$ layer-2-lower-body $>$ trouser $>$ formal-trousers, dressed pants} i.e. \textit{formal-trousers} is synonymous to \textit{dressed pants} and is a type of \textit{trouser} which is again a type of \textit{layer-2-lower body} apparel. Each taxonomy entry has a synonym-set (called ``synset''). With the help of domain experts we collected 282 such fashion ``synsets'' for men and 434 for women \\
%(e.g. \textit{formal-trousers}, \textit{suit pants}, \textit{dressed pants}, \textit{trouser} are fashion ``synsets'')\\
\-\ \-\ \textbf{2b.} Identifying the set of fashion attributes relevant (especially for the purpose of shopping) to each of the fashion synsets. Overall 52 such attributes (like color, pattern, style, price, wash-care information) were identified by domain experts, where 45 of them are visual attributes and remaining are meta-data attributes about the synset (e.g. wash-care information, price, seller ranking)\\
\-\ \-\ \textbf{2c.} Seeding the attribute lexicons with a set of realistic values provided by the domain experts\\
\textbf{3.}\-\ Parsing the semi-structured catalog into a single unified structure of the form $<$synset, $\{$attribute:$\{$attribute values$\}$$\}$$>$,  $\{\}$ denoting a set\\
\textbf{4.} Constructing a distribution of attributes and values for each of these synsets, from the structured catalog data curated in step 3 and filtering them through a close manual inspection by the domain experts\\
\textbf{5.} From the unstructured product description in the catalog, spotting and extracting style-tip information (e.g. black trousers go well with white shirt)\\
\textbf{6.} Creating fashion profiles for celebrities based on the type of clothes and accessories worn or endorsed by them. 
%Since obtaining a large number of such celebrity fashion profile is hard and the data available online is also very noisy 
Since the profile information for real celebrities was proprietary, we generated profiles of imaginary celebrities by simulating a distribution of fashion synsets that each of these celebrities endorse, and a further distribution of fashion attributes preferred by these celebrities for each of these synsets. Note that doing so does not affect the generality of the dataset technically. Statistics about the final domain knowledge curated using this semi-automated methodology are tabulated in Table \ref{tab:knowledge_stats}

\begin{table}
{\tiny
\begin{center}
\begin{tabular}{|p{3.1cm}|p{0.7cm}|p{3.3cm}|} \hline
\multicolumn{2}{|c|}{\textbf{Knowledge Base Statistics}} & \textbf{Examples} \\ \hline
\#Items Crawled & 1.05M & -\\\hline
Avg. \#Images per item & 4 & -\\\hline
\#Fashion Synsets& 716 & shirt, trouser, tuxedo, loafer, stilletoes, sunglasses, handbag, hat \\\hline
\#Fashion Attributes & 52 & color, pattern, material, brand, style, sleeves, price, wash-care, \\\hline
\#Visual Fashion Attributes & 45 & color, pattern, material, style, neck, sleeves, length, sole type, closure \\\hline
\#Meta-Info Attributes & 8 & price, wash-care, product ranking, brand, size, occasions \\\hline
Avg. \#Fashion Attribute per Synset & 16 & -\\\hline
Avg. \#values per attribute & 500 & - \\\hline
\#Coarse-Grained StyleTips (Synset, Synset association)& 8871 & shirt \& trouser, tshirt \& sneakers, tuxedo \& cufflinks, suit \& formal shoes, kurta \& jeans\\\hline
\#Fine-Grained StyleTips (Attribute Synset, Attribute Synset association) & 350K & white shirt \& black trousers, light t-shirt \& dark jacket, black gown and silver stilettoes \\ \hline
\#Celebrity profiles & 411 & Celeb1 likes vegan-leather shoes, monochrome pastel shaded t-shirts and polyester jackets \\ \hline
Avg. \#Synsets endorsed by a celebrity & 4 & Celeb1 endorses lehenga, saree, blouse, kurti, sandals, stilettos. \\ \hline
Avg. \#Celebrities endorsing a synset & 15 & - \\ \hline
\#Synsets endorsed by celebrities & 252 & -\\ \hline
\end{tabular}
\end{center}
}
\caption{Domain Specific Knowledge Base Statistics}
\label{tab:knowledge_stats}
\end{table}

\subsubsection{Gathering multimodal dialogs}
% \subsubsection{Automata to Simulate dialogs}

During the interviews, the domain experts described in detail various phases of the sales process. For example, a dialog between the sales agent and a shopper who visits an e-commerce website with the objective of either buying or browsing one or more fashion items begins by the shopper providing their shopping requirements to the agent. The agent then browses the corpus and comes back with a multimodal response (i.e. with a set of images that satisfy the shopper's constraints and/or some associated text). Now, using this response the shopper provides feedback or modifies their requirements. Through this iterative response and feedback loop the shopper continues to explore their items of interest, adding chosen items to their shopping cart. The session continues until they either choose to exit without a purchase or culminates with the shopper buying one or more items. Note that at various steps during such a conversation, the response of the current step of the dialog is based on inference drawn from an aggregate of images and text in the unstructured dialog context as well as a structured background domain knowledge (which is again multimodal in nature).

\begin{table}[t]
{%\renewcommand{\arraystretch}{0.89}% for the vertical padding
{\scriptsize
\captionsetup{font=small}
\begin{center}
\begin{tabular}{|p{0.3cm}|p{1.3cm}|p{5.5cm}|}
\hline
\textbf{Id}&\textbf{Automata State Type} & \textbf{State Description}\\ \hline
1 &\textbf{greeting} & Shopper greets \\ \Xhline{2\arrayrulewidth} 
2 &\textbf{self-info} & Shopper gives information about him/herself \\  \Xhline{2\arrayrulewidth} 
3 & \textbf{give-criteria} &  describes his requirements\\  \Xhline{2\arrayrulewidth} 
4 &\textbf{show-image} & Agent shows relevant responses for the Shopper's query\\  \Xhline{2\arrayrulewidth} 
5 & \textbf{give-image-description} & Agent generates short description of the product, using visual and catalog information\\ \Xhline{2\arrayrulewidth} 
6 & \textbf{Like/Dislike specific items / overall items, show-more} & 
Shopper expresses negative or positive preference specifically towards one or more items previously or currently shown, or a overall general preference towards all the items and optionally shows a new image to possibly modify his requirements and wants to see more \\ \Xhline{2\arrayrulewidth} 
%\textbf{like previous item(s), show-more} & \\  \Xhline{2\arrayrulewidth} 
%\textbf{like/dislike current items, show-more} & \\ \cline{1-1}
%\textbf{like specific items, show more} & \\  \Xhline{2\arrayrulewidth} 
%\textbf{Like/Dislike overall items, show more} & \\ \hline
%\multirow{1}{*}{\parbox{5.7cm}{ 
%Shopper expresses a general preference (negative or positive) towards the currently displayed items, and optionally shows a new image to possibly modify his requirements and wants to see more\\ \hline
%\textbf{like overall items, show more} & \\  \Xhline{2\arrayrulewidth} 
7 & \textbf{show-orientation} & Shopper wants to see an item from different orientations\\  \Xhline{2\arrayrulewidth} 
8 & \textbf{show-similar} & Shopper wants to see similar to a particular item\\  \Xhline{2\arrayrulewidth} 
9 & \textbf{goes-with} & Shopper asks for style-tip\\  \Xhline{2\arrayrulewidth} 
10 & \textbf{ask-attribute} & Shopper asks about the attributes of the items shown\\  \Xhline{2\arrayrulewidth} 
11 & \textbf{suited-for} & Shopper asks about what are suited for that item\\  \Xhline{2\arrayrulewidth} 
12 & \textbf{sort-results} & Shopper wants to sort the result set by some attribute\\  \Xhline{2\arrayrulewidth} 
13 & \textbf{filter-results} & Shopper wants to filter the results based on some attribute\\  \Xhline{2\arrayrulewidth} 
14 & \textbf{celebrity} & Shopper asks questions relating to some celebrities and his fashion items of interest\\  \Xhline{2\arrayrulewidth} 
15 & \textbf{switch-synset} & Shopper wants to switch back to the type of fashion synset he had seen previously\\  \Xhline{2\arrayrulewidth} 
16 & \textbf{buy} & Shopper wants to buy one or more items\\  \Xhline{2\arrayrulewidth} 
17 & \textbf{exit} & Shopper wants to exit \\ \hline
\end{tabular}
\end{center}
}
}
\caption{Details of example Automata State-Types as described by domain experts. 14 of these state-types in turn have multiple different states, thus yielding overall 84 automata states to be used in the dialog}
\label{tab:automatastates}
\end{table}

The domain experts described each of these various possible types of states involved in the conversations, a subset of which are shown in Table \ref{tab:automatastates}. This was mapped to an expert model automata which consisted of a total of 17 state types for the shopper covering 84 states. Each such state had a specific intent and corresponding to them, a collection of representative utterance patterns involving both text and images were designed by us along with the experts. Each such state would exhibit the following 3 features: (a) multimodality of utterance/response: shopper's utterance and the agent's response could involve either text or image or both, (b) multimodality of context: shopper's utterance would employ the context of the conversation which would include both the text history and a number of images and (c) combination of the structured domain knowledge and the unstructured dialog context, both being multimodal.

The domain experts then provided a large number of possible sales process flows of customers proceeding from one state to another. These transitions were captured by the automata between these states with expert designed transition probabilities. The domain experts then inspected the outputs of small runs of the automata and provided feedback. This feedback was then incorporated to further refine the expert automata whose runs were again inspected by the experts. This iterative process was manually-intensive, and required close coordination with the domain experts. Following this process, we produced a large-scale dataset of over 150K multimodal dialogs.

\begin{table}
{\tiny
\captionsetup{font=small}
\begin{center}
\begin{tabular}{|p{4.2cm}|p{.75cm}|p{.75cm}|p{.75cm}|}\hline
\textbf{Dataset Statistics} & \textbf{Train} & \textbf{Valid} & \textbf{Test} \\ \hline
\#Dialogs(chat sessions) & 105,439 & 22,595 & 22,595 \\ \hline
Proportion in terms of dialogs & 70\% & 15\% & 15\% \\ \hline
Avg. \#Utterances per dialog & 40 & 40 & 40\\\hline
\#Utterances with shopper's questions & 2M & 446K & 445K \\\hline
\#Utterances with agent's image response & 904K & 194K & 193K \\\hline
\#Utterances with agent's text response & 1.54M &331K & 330K \\\hline
Avg. \#Positive images in agent's image response & 4 & 4 & 4 \\\hline
Avg. \#Negative images in agent's image response & 4 & 4 & 4\\\hline
%No. of pairs of positive-negative images in agent's image response & 1304,721 & 2813,861 & 2804,062\\\hline
Avg. \#Words in shopper's Question & 12 & 12 & 12\\\hline
Avg. \#Words in agent's text Response & 14 & 14 & 14\\\hline
Avg. \#Automata states per dialog & 15 & 15 & 15\\\hline
Avg. \#Dialogs having a particular automata state & 20,246 & 4,346 & 4,335\\ \hline
Avg. \#Automata state-types per dialog & 13 & 13 & 13\\\hline
Avg. \#dialogs having a particular state-type & 59,638 & 12,806 & 12,764\\\hline
\#Automata states & 84 & 84  & 84 \\\hline
\#Automata state types & 22 & 22 & 22 \\\hline
Vocabulary Size (threshold frequency$>$=4) & 26,422 & - & -\\\hline
\end{tabular}
\end{center}
}
\caption{Multimodal Dialog Dataset Statistics}
\label{tab:dataset_stats}
\end{table}

\begin{table*}[ht]
{\tiny
\begin{center}
\begin{tabular}{|p{2.5cm}|p{8.2cm}|p{5.2cm}|}\hline
\textbf{Type of Complexity} & \textbf{Example State} & \textbf{Example Utterance} \\ \hline
\textbf{Long-Term Context} & At the beginning of the dialog the user mentions his budget or size preference and after a few utterances, asks the agent to show something under his budget or size & \textit{I like the 4th image. Show me something like it but in style as in this image within my budget.}\\\hline
\textbf{Quantitative Inference ~~~(Counting)} & User points to the $n$th item displayed and asks a question about it & \textit{Show me more images of the 3rd product in some different directions}\\\hline
\textbf{Quantitative Inference ~~~(Sorting / Filtering)} & User wants sorting/filtering of a list based on a numerical field, e.g. price or product rating & \textit{Show me ... within my budget.}\\\hline
\textbf{Logical Inference} & User likes one fashion attribute of the $n$th image displayed but does not like another attribute of the same & \textit{I am keen on seeing something similar to the 1st image but in a different sole material} \\\hline
\textbf{Visual Inference} & System adds a visual description of the product alongside the images & \textit{Viscata shoes are lightweight and made of natural jute, premium leather, suedes and woven cloth}\\\hline 
\textbf{Inference over aggregate of Images} & User's question can have multiple aspects, drawn from multiple images displayed in the current or past context & \textit{List more in the upper material of the 5th image and style as the 3rd and the 5th}\\\hline
\textbf{Multimodal Inference} & User gives partial information in form of images and text in the context & \textit{See the first espadrille. I wish to see more like it but in a silver colored type}\\\hline
\textbf{Inference using domain knowledge and context} & Sometimes inferences for the user's questions go beyond the dialog context to understanding the domain & \textit{Will the 5th result go well with a large sized messenger bag?}\\\hline
\textbf{Coreference Resolution / Incomplete Question} & Temporal continuity between successive questions from the user may cause some of them to be incomplete or to refer to items or aspects mentioned previously & \textit{Show me the 3rd product in some different directions ... What about the product in the 5th image?}\\\hline
\end{tabular}
\end{center}
}
\caption{Anecdotal examples of different aspects of complexity in the MMD dataset. Examples in the $3^{rd}$ column are snippets of the dialog session illustrated in Figure\ref{fig:dialogue}}
\end{table*}

\begin{table}[!htbp]
{\tiny
\captionsetup{font=small}
\begin{center}
\begin{tabular}{|p{1.84cm}|p{0.7cm}|p{4cm}|}\hline
\textbf{\% of Surveyed 760 Dialogs} & \textbf{Rating} & \textbf{Rating Chart} \\ \hline
~~~~~~~42.0\% & ~~5 & dialog is realistic with no errors \\ \hline
~~~~~~~30.0\% & ~~4 & $<=$2 Conversational Errors\\ \hline
~~~~~~~19.2\% & ~~3 & $<=$2 Conversational $<=$2 Logical Errors  \\ \hline
~~~~~~~6.8\% & ~~2 & $<=$2 Conversational $<=$4 Logical Errors \\ \hline
~~~~~~~2.0\% & ~~1 & $>$2 Conversational and $>$4 Logical Errors \\ \hline
\end{tabular}
\end{center}
}
\caption{Domain expert ratings in the qualitative survey.}
\label{tab:survey}
\end{table}

\subsubsection{Qualitative Survey}

To ensure that the dataset is representative and not biased by the specific fashion experts interviewed, we conducted a survey of the dataset by involving a different set of 16 fashion experts. They were asked to evaluate both whether the text portions of the dialog are natural sounding and meaningful and whether the images in it are appropriate. The survey was conducted with a randomly sampled set of 760 dialog sessions and the experts were asked to provide an overall rating between 1 to 5 (with 5 being most realistic). 

Two types of errors were documented: (i) minor error being conversational mistakes (e.g. grammatical and phrasing error), (ii) severe error being logical mistakes (e.g. deductions errors in generating the image or text response, incorrect understanding of the shopper's question, wrong fashion recommendation, etc.). As the survey results in Table \ref{tab:survey} show, the average rating obtained was 4 out of 5, thereby implying that on average there were only a few conversational errors in a typical sized dialog session consisting of about 40 utterances.

Of course, the dataset still contains some noise which is inherited either from the noise in the original catalogs crawled from various websites or because of the process used for creating structured data from unstructured textual descriptions. For example, a product titled ``California Bear Logo Flag Republic Flats Bill Snapback'' is actually a type of ``cap'' but when populating the structured data it was wrongly labeled as a ``shoe'' because ``flats'' is a valid shoe-type. While such errors exist, they are very minimal and do not affect the overall quality of the dataset. Such noise is expected in any dataset created at this scale. Our manual survey suggested that most dialogs in the dataset have very few (if any) such logical errors.

\subsection{Tasks}

The proposed MMD dataset consists of multimodal, domain-aware conversations between 2 agents, and hence can be used for evaluating a wide variety of tasks. We describe each of these tasks and explain the technical challenges involved:\\
\noindent\textbf{1. \emph{Text Response}}: Given a context of $k$ turns the task here is to generate the next text response.\\
\textbf{2. \emph{Image Response}}: Given a context of $k$ turns the task here is to output the most relevant image(s). There are 2 typical approaches to achieve this:\\
\textbf{2.1 \emph{Image Retrieval}}: Given a context of $k$ turns and a database of images, retrieve and rank $m$ images based on their relevance to the given context. \\
\textbf{2.2 \emph{Image Generation}}: Given a context of $k$ turns, generate the most relevant image (typically performed using generative models e.g. contextual GANs\cite{reed2016generative,DBLP:conf/nips/GoodfellowPMXWOCB14}).\\
We propose both tasks since the evaluation criteria for each approach is quite different.\\
\textbf{3. \emph{Employing Domain-knowledge}}: This is essentially performing tasks (1) and (2) of text and image response generation using both the unstructured dialog context along with the structured domain knowledge. We propose this as a separate task to evaluate the impact of domain-knowledge.\\
\textbf{4. \emph{User Modeling}}: Another important conversation aspect is to study the varying shopping behavior of users e.g. their buying preferences, speed of decision making, etc. Hence, we propose a task to explicitly model the shopper since it impacts the agent's most appropriate response at each step.
\\\\
\noindent \textbf{Setup:} In this work, we focus on tasks (1) and (2.1) and make two simplifications: (a) We evaluate the text response and image response task separately, which means that the system does not need to decide the modality of the response, and (b) instead of retrieving and ranking all of the images in the catalog/database, the system needs to rank only a given smaller subset of $m$ images, which contain the correct image(s) in addition to a few incorrect ones. This simplified evaluation protocol of ``selecting/ranking'' the best response was proposed for the Ubuntu Dialog Corpus \cite{DBLP:conf/sigdial/LowePSP15} and helps provide more control on the experimental setup and evaluate individual parts of the system in a more thorough manner.

\subsection{Dataset Versions}
\label{subsec:dataset_versions}
To validate the text generation and image selection tasks in this work, we create two datasets\\
\noindent\textbf{Version 1}: This version includes the ``give-image-description'' state (as in Table \ref{tab:automatastates}) where the system may also provide a short description about the images, while displaying them. For example, in Figure\ref{fig:dialogue} the system gives a crisp product description ``\emph{michael kors brings you an effortless weekend look with the wrapped espadrille heel}'' along with the images\\
\noindent\textbf{Version 2}: This version of the dataset is exactly identical to the first version, except that the utterances corresponding to the ``give-image-description'' state are missing in this version, i.e. the system will not provide short descriptions of a product, upon displaying its image. The remaining utterances and the flow of the dialogs are identical in both the versions.

It should be noted that the first version poses a harder challenge for the text task, while making the image task somewhat easier, and the second one is more challenging for the image task while being simpler for the text task.

Given this setup, we now propose baseline models for these tasks based on the encode-attend-decode paradigm.

%\subsection{Dataset Noise}
%{\color{red} As the original fashion data had been crawled from different retail sites and parsed through various semi-automated means to get a structured multimodal catalog with product images and attribute-value tags, there is a certain amount of noise in the constructed catalog itself. This impacts the following steps of semi-automated style-tip mining and the final dialog generation.

%In the light of this, it should be noted that the objective of the dataset is not to build a perfect conversation system for fashion browsing but to enable models to handle multimodal data in a conversation and also leverage a mix of unstructured dialog context as well a structured domain knowledge and catalog, (all being multimodal in nature) in responding to an user's query. }

\section{Models}
\begin{figure*}[!htb]
\centering
{
  \includegraphics[ height=5cm,width=0.82\textwidth]{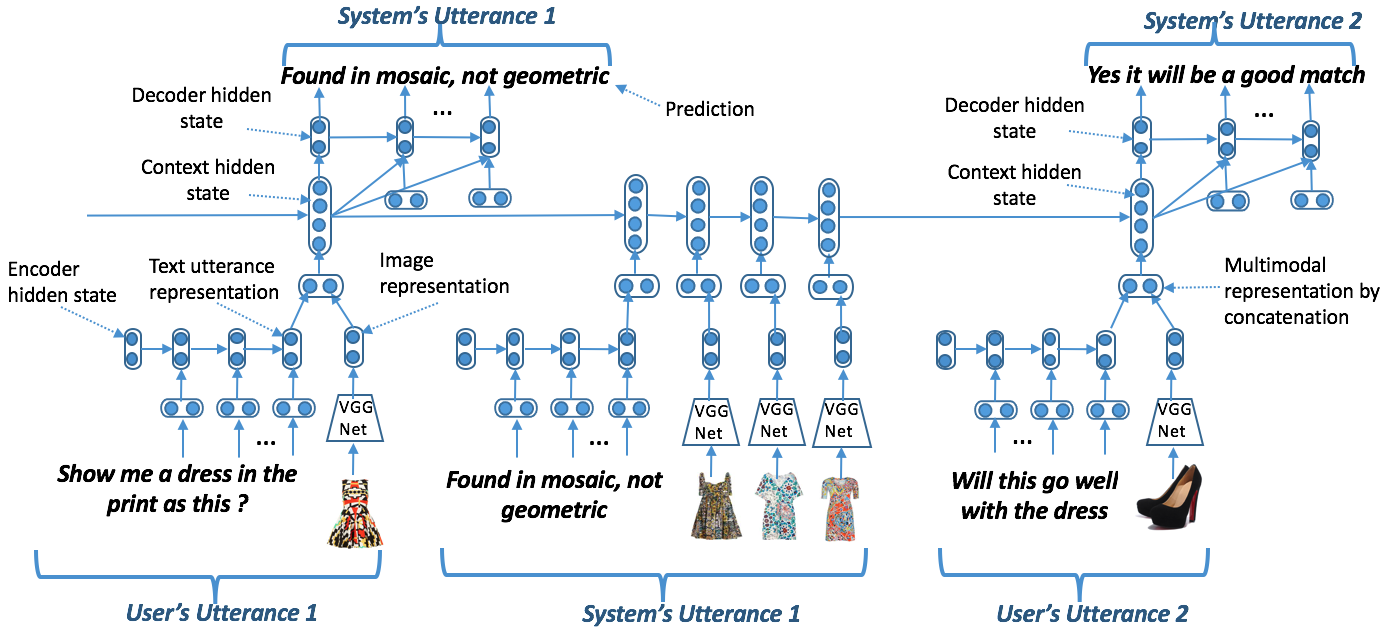}
   \caption{\label{fig:modelTextTask} Multimodal Hierarchical Encoder Decoder Architecture for the Text Response Task. The figure shows all multimodal utterances, but in general utterances can have either modality or both. }
  } 
   
\end{figure*}
\begin{figure*}[htb]
\centering
{
  \includegraphics[height=5.4cm,width=0.89\textwidth]{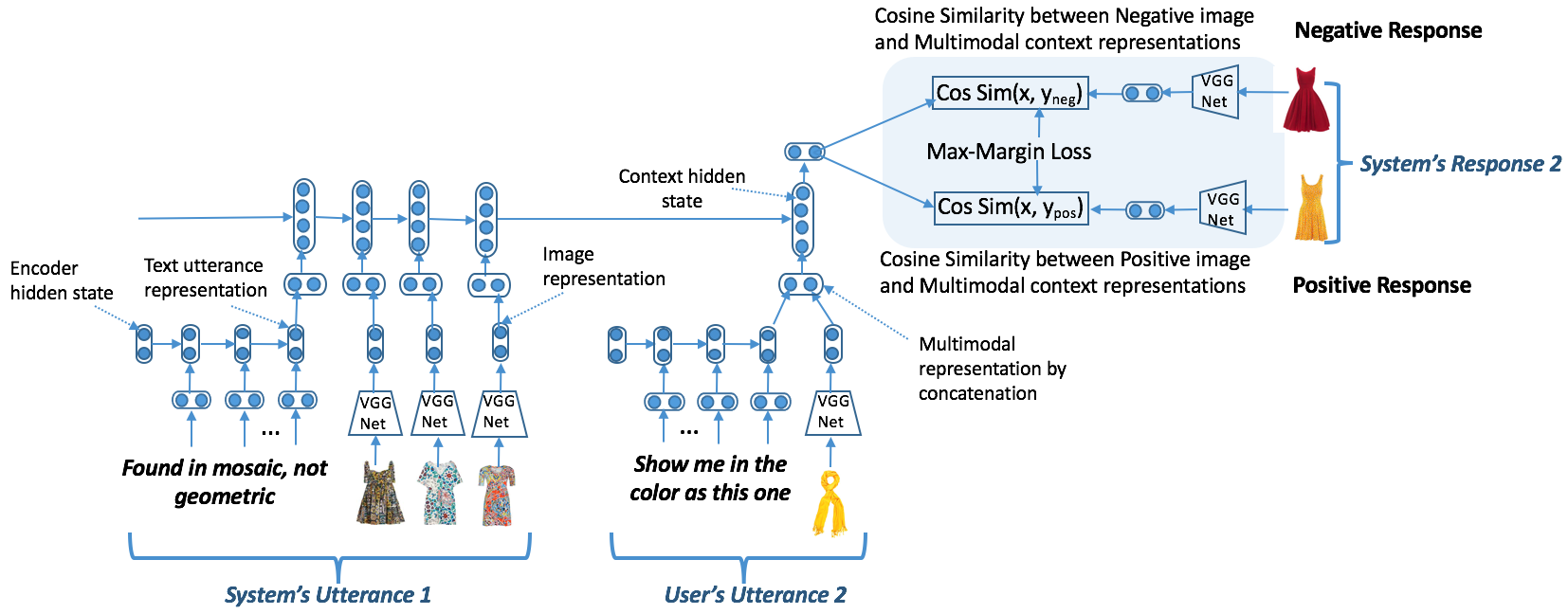}
   \caption{ \label{fig:modelImageTask} Multimodal Hierarchical Encoder Architecture for Image Response Task. The figure shows a single target positive and negative image, but in general, $m$($>$1) images can be provided as target}
 }
  
\end{figure*}
\label{sec:models}
To empirically estimate the feasibility of the tasks described earlier we implement one baseline method (and some variations thereof) for each task, based on the popular hierarchical encode-attend-decode paradigm \cite{Serban:2016:BED:3016387.3016435} typically used for (text) conversation systems. We split the description below into two parts (i) Multimodal encoder which is common for the two tasks (ii) Multimodal decoder which is different depending on whether we need to generate a text response or predict an image response. %Note that even though the encoder is same for the two tasks we do not train it jointly but separately for each task.  

\subsection{Mutimodal encoder}

\iffalse
\begin{figure*}[!htb]
\centering
{
  \includegraphics[ height=5.7cm,width=\textwidth]{images/modeltextimage.png}
   \caption{\label{fig:model} Multimodal Hierarchical Encoder Decoder Architecture for (a) Text Response Task and (b) Image Response Task using the Multimodal Dialog context (elaborated in the leftmost figure). Though the image response task shows a single target positive and negative image, but in general, $m$($>$1) images can be provided as target}
  } 
\end{figure*}
% \begin{figure*}[!htb]
% \centering
% {
%   \includegraphics[height=4.8cm,width=0.87\textwidth]{images/modeltaskimage.png}
%   \caption{ \label{fig:modelImageTask} Multimodal Hierarchical Encoder Architecture for Image Response Task. The figure shows a single target positive and negative image, but in general, $m$($>$1) images can be provided as target}
%  }
  
% \end{figure*}
\fi 
As mentioned earlier, for both the tasks, the context contains $k$ utterances where each utterance could either be (i) a text only or (ii) an image only utterance or (iii) multimodal, containing both text and images (as shown in Figure \ref{fig:modelTextTask} and Fig. \ref{fig:modelImageTask}). In each of these, we use a multimodal hierarchical encoder for encoding the input representation as below.

\noindent \textbf{(a) Text only utterance:} Every text utterance in the context is encoded using a bidirectional RNN network with GRU \cite{chung2014empirical} cells in a process similar to the utterance level encoder described in \cite{Serban:2016:BED:3016387.3016435}. This is the level 1 encoder in the hierarchical encoder.

\noindent \textbf{(b) Image only utterance:} If an utterance contains one or more images, we encode each image using a 4096 dimensional representation obtained from the FC6 layer of a VGGNet-16 \cite{Simonyan14c} convolutional neural network. If an utterance contains multiple images, it is unrolled into a sequence of utterances, with each containing a single image and/or a single text sequence. This is also a part of the first level in the hierarchical encoder.

\noindent \textbf{(c) Multimodal utterance:} The text portion of the multimodal utterance is encoded using the same GRU cells as used for encoding the text only utterance. Similarly, the images in the multimodal utterance are encoded using the same VGGNet-16 as used for the image only utterance. The final representation of the multimodal utterance is simply the concatenation of the individual utterances.

The multimodal utterance representation is then fed to a level two encoder which is again a GRU. This second level (or context-level encoder) essentially encodes the sequence of utterances where the representation of each utterance in the sequence is computed and projected as described above. Fig. \ref{fig:modelTextTask} and Fig. \ref{fig:modelImageTask} shows this process of computing the encoder representation for a given multimodal context.

\subsection{Decoder for generating text responses}
As shown in Fig. \ref{fig:modelTextTask}, we use a standard recurrent neural network based decoder with GRU cells. Such a decoder has been used successfully for various natural language generation tasks including text conversation systems \cite{DBLP:journals/corr/SerbanSLCPCB16}. We also implemented a version where we couple the decoder with an attention model which learns to attend upon different time-steps of the second level encoder (again this has been tried successfully in the context of text conversation systems \cite{DBLP:journals/corr/YaoPZW16}).

\subsection{Layer for ranking image responses:} The task here is to rank a given set of images depending on their relevance to the context. While training we are given a set of $m$ images for each context of which only $n_{pos\_max}$ are relevant for the context. The remaining $m - n_{pos\_max}$ are picked from the corresponding false image responses in the dataset. 
%If that utterance in the datset has insufficient number of false responses, the remaining are picked randomly from the entire collection of images. 
We train the model using a max margin loss. Specifically, we compute the cosine similarity between the learnt image embedding and the encoded representation of the multimodal context. The model is then trained to maximize the margin between the cosine similarity for the correct and the incorrect images. Fig. \ref{fig:modelImageTask} depicts this for the case when $m = 2$ and $n_{pos\_max} = 1$. Due to space constraints we only provide the model's pictorial representations.

\section{Experiments}
\label{Sec:Experiments}
\begin{table}[!ht]
\centering
{%\renewcommand{\arraystretch}{0.89}% for the vertical padding
{\scriptsize
\captionsetup{font=small}
\begin{tabular}{|m{0.4cm}|m{2.1cm}|m{0.55cm}|m{0.55cm}|m{0.55cm}|m{0.55cm}|m{0.55cm}|}\hline
       \textbf{Ver.} & \textbf{Model} & \multicolumn{2}{c|}{\textbf{Text Task}} & \multicolumn{3}{c|}{\textbf{Image Task ($m$=5)}} \\ \cline{3-7}
     &\textbf{(with Context Size)} & \textbf{Bleu} & \textbf{Nist} & \textbf{R@1} & \textbf{R@2} & \textbf{R@3} \\ \hline
    \multirow{ 5}{*}{V1} & Text-Only HRED (5) & 14.58 & 2.61 & 0.46 & 0.64 & 0.75\\ \cline{2-7}
    & \textbf{Multimodal HRED} (2) & \textbf{20.42} & \textbf{3.09} & 0.72 & 0.86 & 0.92 \\ \cline{2-7}
    & Multimodal HRED (5) & 19.73 & 2.94 & 0.71 & 0.86 & 0.92\\ \cline{2-7}
    & \textbf{Attention based Multimodal HRED (2)} & 19.58 & 2.46 & \textbf{0.79} & \textbf{0.88} & \textbf{0.93} \\ \cline{2-7}
    & Attention based Multimodal HRED (5) & 19.37 &2.51 & 0.68 & 0.84 & 0.91\\ \hline\hline
    
    \multirow{ 8}{*}{V2} & Text-Only HRED (5) & 35.9 & 5.14 & 0.44 & 0.6 & 0.72\\ \cline{2-7}
    %& Multimodal HRED with concatenated image & 2 & 32.99 & 4.83 & 0.43 & 0.6 & 0.73\\ \cline{2-8}
    %& Multimodal HRED with concatenated image & 5 & 34.77 & 5.05 & 0.44 & 0.61 & 0.73\\ \cline{2-8}
    %& Multimodal HRED with concatenated image & 10 & 37.43 & 5.24 & 0.37 & 0.55 & 0.69\\ \cline{2-8}
    & \textbf{Multimodal HRED} (2) & \textbf{56.67} & \textbf{7.51} & 0.69 & 0.85 & 0.90\\ \cline{2-7}
    & Multimodal HRED (5) & 56.15 & 7.27 & 0.68 & 0.84 & 0.90\\ \cline{2-7}
    & \textbf{Attention based Multimodal HRED (2)} & 50.20 & 6.64 & \textbf{0.78} & \textbf{0.87} & \textbf{0.923}\\ \cline{2-7}
    & Attention based Multimodal HRED (5) & 54.58 & 6.92 & 0.66 & 0.83 & 0.89\\ \hline
\end{tabular}
}
}
\caption{\label{tab:results} Performance of the different models on the  ``Text Response Generation'' and ``Best Image Selection'' Task.}

\end{table}

\begin{table}[!ht]
\centering
{\scriptsize
%\begin{tabular}{|>{\raggedright}p{0.8cm}|>{\raggedright}p{0.18cm}|>{\raggedright}p{0.39cm}|>{\raggedright}p{0.39cm}|>{\raggedright}p{0.39cm}|>{\raggedright}p{0.39cm}|>{\raggedright}p{0.39cm}|>{\raggedright}p{0.39cm}|>{\raggedright}p{0.39cm}|>{\raggedright}p{0.39cm}|}\hline
\begin{tabular}{|p{0.23cm}|p{0.38cm}|p{0.45cm}|p{0.42cm}|p{0.42cm}|p{0.42cm}|p{0.42cm}|p{0.423cm}|p{0.42cm}|p{0.42cm}|}\hline
 \textbf{Id}& \textbf{Ver} & \multicolumn{2}{c|}{\textbf{Text Task}} & \multicolumn{6}{c|}{\textbf{Image Task}} \\ \cline{5-10}
  & &  \multicolumn{2}{c|}{\textbf{}} &  \multicolumn{3}{c|}{\textbf{$m$=50}} &  \multicolumn{3}{c|}{\textbf{$m$=100}} \\ \cline{3-10}
  
 & & Bleu & Nist & r@1 & r@2 & r@3 & r@1 & r@2 & r@3 \\ \hline

\multirow{3}{0.23cm}{10} 
& V1&	44.42	& 5.92		& -&	-&	-& - &	 - &	-\\ \cline{2-10}
 & V2 & 45.21 &	6.04	& 	 - &	 - &	-& - &	 - &	-\\ 
 & & & & & & & & &  \\ \hline

%\multirow{2}{1.8cm}{buy} 
%& V1 &13.6K &	3.88 &	1.56 &	 \multirow{2}{*}{0}	& -&	-&	-& - &	 - &	-& - &	 - &	-\\ \cline{2-5} \cline{7-15}
%& V2 & 13.6K	& 7.23 &	1.35 &		&-&	-&	-& - &	 - &	-& - &	 - &	-\\ \hline 

\multirow{2}{0.23cm}{14} 
& V1 	& 5.79 &	0.662 &	-&	-&	-& - &	 - &	-\\ \cline{2-10}
 & V2 	& 13.65 &	1.327 &	 - &	 - &	-& - &	 - &	-\\
   \Xhline{2.7\arrayrulewidth}  

\multirow{2}{0.23cm}{9}
& V1 	&28.37	&1.164 &		-	&-	&-& - &	 - &	-\\ \cline{2-10}
 & V2  & 	31.47 &	2.746 &		& - &	 - &	-& - &	-\\  
 & & & & & & & & &  \\ \Xhline{2.7\arrayrulewidth}  

\multirow{2}{0.23cm}{11} 
 & V1	& 67.2	&6.23&	-	&-&	-& - &	 - &	-\\ \cline{2-10}
  & V2  &	60.92	& 5.54 &		& - &	 - &	-& - &	-\\  
 & & & & & & & & &  \\ \Xhline{2.7\arrayrulewidth}

 \multirow{2}{0.23cm}{6} & V1 	& 4.18	& 0.697 & 0.18 & 0.32 & 0.43	 & 0.11 & 0.20 & 0.27\\ \cline{2-10}
 & V2 	& 80.56 &	8.20 & 0.225 & 0.28 & 0.41 & 0.078 & 0.127 & 0.179 	\\  
 & & & & & & & & &  \\ \Xhline{2.7\arrayrulewidth}

\multirow{2}{0.23cm}{13} 
& V1	&19.16	& 1.92 & 0.24 & 0.38 & 0.49 & 0.15 & 0.25 &0.34 \\ \cline{2-10}
& V2 &	86.57&	3.98 & 0.176 & 0.298 & 0.367 & 0.083 & 0.15 & 0.21\\  
 & & & & & & & & &  \\ \Xhline{2.7\arrayrulewidth}  
 
\multirow{2}{0.23cm}{7}
 & V1 & 	99.79 &	6.36 & 0.26 & 0.40 & 0.50 &0.13 &0.23 &0.31	\\ \cline{2-10}
  & V2 &  99.98 &	6.37 &0.21 & 0.336 & 0.43 &  0.1197 & 0.204 & 0.27	\\  
 & & & & & & & & &  \\ \Xhline{2.7\arrayrulewidth}  
 
\multirow{2}{0.23cm}{8}
 & V1 & 	99.87	& 2.32	& 0.23 & 0.37 & 0.48 & 0.15 & 0.25 & 0.33\\ \cline{2-10}
  & V2 	& 100 & 	2.32 & 0.17 & 0.289 & 0.383 & 0.10 & 0.177 & 0.240 \\  
 & & & & & & & & &  \\ \Xhline{2.7\arrayrulewidth}  
 
\multirow{2}{0.23cm}{12} 
& V1	&100	&5.21	 & 0.24 & 0.38 & 0.49 & 0.09 & 0.18 & 0.26\\ \cline{2-10}
& V2 	& 100 &	5.21 &0.136 & 0.229 & 0.31 & 0.084 & 0.144 & 0.194  \\  
 & & & & & & & & &  \\ \hline 

 \end{tabular}
}
\caption{\label{tab:statewiseresults} Best Model's performance on dialog states described in Table \ref{tab:automatastates} (`Id' refers to the ID of the state-type in Table \ref{tab:automatastates}) and V1, V2 refer to the two versions of the dataset, $m$ refers to the size of target image set (one correct, rest incorrect) to be ranked by the model and is varied from $50$ to $100$} 
\end{table}

Now we describe the experimental setup used to evaluate the following models on the two tasks:
\begin{enumerate}[$\bullet$,leftmargin=*,topsep=0pt]
\item \textbf{Hierarchical Encoder Decoder(HRED) (ignoring image context)}, whose architecture is similar to that proposed in \cite{Serban:2016:BED:3016387.3016435} %and acts as a simple baseline 
\item The proposed \textbf{Multimodal Hierarchical Encoder Decoder(HRED)}, (both with and without attention)
\end{enumerate}

\subsection{Evaluating the Text Response Task}
For this task we only considered those dialog turns ending with a text response. The training, validation and test sets sizes are reported in the 6$^{th}$ row of Table \ref{tab:dataset_stats}. We used Adam  optimization algorithm and tuned the following hyperparameters using the validation set; learning rate $\in$ \{1e-3, 4e-4\}, RNN hidden unit size $\in$ \{256, 512\}, text and image embedding size $\in$ \{256, 512\}, batch size $\in$ \{32, 64\} and dialog context size $\in$ \{2,5,10\}. The bracketed numbers indicate the values of each hyperparameter considered. Table \ref{tab:results} summarizes the BLEU and NIST %(\url{ftp://jaguar.ncsl.nist.gov/mt/resources/}) 
scores used for evaluation.% and we discuss them in the next section. 

\subsection{Evaluating Image Response Task}

During training and evaluation for this task we only consider those dialog turns ending in an image response from the system. The training, validation and test sets sizes are reported in the 5$^{th}$ row of Table \ref{tab:dataset_stats}. Both during training and testing, the model is provided with $m$=5 target images out of which only $n_{pos\_max}$=1 is relevant and at test time the model has to rank the images in order of their relevance as a response to the given context. The hyperparameters of the model were tuned in the same way as mentioned above. Note that for evaluating the image response in selection/ranking mode, a system would also need negative training and test examples alongside the correct ones. Negative examples are generated by either sampling an item from the wrong fashion category (e.g. shoe in place of a bag) or a wrong sub-category of the target item (e.g a backpack in place of a sachel bag) or items violating certain attribute criteria provided by the user.
\\
\-\ \-\ We use \textit{Recall@top-m} as the evaluation metrics where \textit{top-m} is varied from 1 to 3, and the model prediction is considered to be correct only if the true response is among the \textit{top-m} entries in the ranked list. These are summarized in Table \ref{tab:results}.

%For Dataset2, the text performance is reasonably well for most of the states. However for the image response task, the variance over the different states is much less pronounced. This showcases the artifact about the negative examples; (as discussed in Section \ref{subsec:discussion}) that they are not very sensitive towards some of the dialog states but overall are bound to the main conversation thread. However, the per-state evaluation of Image response retrieval task (which is one of the future research challenges brought out by this dataset), would bring out the main challenges of a realistic system.

\subsection{Discussions}
We make a few observations from the results. % in Table \ref{tab:text_results} and \ref{tab:image_results}.

\begin{enumerate}[$\bullet$,leftmargin=*,topsep=0pt]
\item For both tasks, the ``Multimodal HRED model with image sequence'' performs significantly better than both the unimodal baseline HRED models thus suggesting that adding images indeed improves inference for both the tasks. 
\item Comparing the performance of the text response tasks on the two dataset versions, it is obvious that the model performs fairly well on all kinds of text responses except the ``give-image-description'' response, which in itself is a very hard task as it exploits both visual features and domain knowledge or other catalog information.
\item Further, comparing the image response performance for the two dataset version, we observe that having additional textual descriptions of the images in a dialog context can help in better image response selection, which is intuitive.
\item Adding attention does not improve performance. Though counter intuitive, this suggests the need for better multimodal attention models.
\item In Table \ref{tab:statewiseresults}, we have reported the performance of the best model trained for the Text and Image Response Task (as per Table \ref{tab:results}) on the 9 most significant and frequently occurring user-initiated dialog states described earlier in Table \ref{tab:automatastates}. As is evident from the table, the text response performance shows a high variance over the dialog states especially for Dataset V1, thus indicating that wherever the system needs to respond with a short product description, requiring core domain knowledge, it performs poorly.
\item Further, in Table \ref{tab:statewiseresults}, we also report the performance on the Image Task obtained by varying $m$. When we use 50 or 100 candidate images (instead of 5) we see a sharp decline in the performance, indicating that a better model is needed to retrieve and rank images from a large corpus of images. 
\item Overall, we feel there is enough scope for improvement and the current models only establish the feasibility of the two tasks. Further, we believe that benchmarking the performance on salient dialog states will allow for more focused research by understanding the challenges and complexities involved in each of these states. 
\end{enumerate}

%\subsection{Per-State Evaluation of the Text and Image Response Task}
 %In certain other states like `ask-attribute'  or `go-with', which can be reasonably answered from the multimodal dialog context itself, the performance is somewhat better. 
%Further, states having very high BLEU score but very low NIST score suggest that the the model performs poorly in predicting the more informative words. 
%\\
%\item In Table \ref{tab:statewiseresults}, we also report the performance on the Image Task obtained by varying $m$. Specifically, we use 50 and 100 candidate images instead of 5. We see a sharp decline in the performance as the number of candidate images increases, indicating that a more sophisticated model is needed to retrieve and rank images from a large corpus of images. Even in the most lenient setting (where $m$=5), the model performs poorly wherever any quantitative inference is required (e.g., `sort-results' or `filter-results'). This highlights the challenges involved in learning such numerical functions. 
%\subsection{Details of the Dataset Released}

To facilitate further research on multimodal systems, the MMD dataset created as a part of this work will be made available at  \url{https://github.com/iitm-nlp-miteshk/AmritaSaha/tree/master/MMD} (please copy paste the URL in a browser instead of clicking on it). This URL will contain the following resources:
\begin{itemize}
\item the train, valid, test splits of the two versions of the MMD dataset and the script to extract the state-wise data for each of the states elaborated in Table \ref{tab:automatastates}
\item the multimodal catalog data in the raw form (before parsing) consisting of unstructured text descriptions and images of the product. But it should be noted that for the current benchmarked models, we only used the image information from the catalog and the multimodal context from the ongoing dialog
\item the domain specific knowledge-base (\textit{i.e.}, the fashion taxonomy, attribute lexicons, style-tips, celebrity profiles \textit{etc.}) curated from the parsed catalog
\end{itemize}

\section{Conclusion}
In this paper, we introduced the Multimodal Dialogs (MMD) dataset curated by working closely with a group of 20 fashion retail experts and consisting of over 150K multimodal conversation sessions between shoppers and sales agents. We proposed 5 new sub-tasks along with their evaluation methodologies. We also showcased 2 multimodal neural baselines using the encode-attend-decode paradigm and demonstrated their performance on both text response generation and best image response selection. Their performance demonstrate the feasibility of the involved sub-tasks and highlight the challenges present. Finally, we suggest new research directions for addressing the challenges in multimodal conversation.

\bibliography{aaai}
\bibliographystyle{aaai}

\end{document}